\documentclass{article}

\usepackage{microtype}
\usepackage{graphicx}
\usepackage{subfigure}
\usepackage{booktabs} 
\usepackage{setspace} 
\usepackage{amsmath}
\DeclareMathOperator*{\argmax}{argmax}
\usepackage{amssymb}

\usepackage{graphicx}

\makeatletter
\newcommand*\bigcdot{\mathpalette\bigcdot@{0.8}}
\newcommand*\bigcdot@[2]{\mathbin{\vcenter{\hbox{\scalebox{#2}{$\m@th#1\bullet$}}}}}
\makeatother

\usepackage{hyperref}
\usepackage{multirow,makecell}

\usepackage[accepted]{icml2021}

\icmltitlerunning{Controllable Missingness from Uncontrollable Missingness}

\begin{document}

\twocolumn[
\icmltitle{Controllable Missingness from Uncontrollable Missingness:\\ Joint Learning Measurement Policy and Imputation}



\icmlsetsymbol{equal}{*}

\begin{icmlauthorlist}
\icmlauthor{Seongwook Yoon}{kor}
\icmlauthor{Jae Hyun Kim}{kor}
\icmlauthor{Heejeong Lim}{kor}
\icmlauthor{Sanghoon Sull}{kor}
\end{icmlauthorlist}

\icmlaffiliation{kor}{Korea University, Seoul, Korea}

\icmlkeywords{Machine Learning}

\vskip 0.3in
]


\printAffiliationsAndNotice{}  

\begin{abstract}
Due to the cost or interference of measurement, we need to control measurement system. Assuming that each variable can be measured sequentially, there exists optimal policy choosing next measurement for the former observations. Though optimal measurement policy is actually dependent on the goal of measurement, we mainly focus on retrieving complete data, so called as imputation. Also, we adapt the imputation method to missingness varying with measurement policy. However, learning measurement policy and imputation requires complete data which is impossible to be observed, unfortunately. To tackle this problem, we propose a data generation method and joint learning algorithm. The main idea is that 1) the data generation method is inherited by imputation method, and 2) the adaptation of imputation encourages measurement policy to learn more than individual learning. We implemented some variations of proposed algorithm for two different datasets and various missing rates. From the experimental results, we demonstrate that our algorithm is generally applicable and outperforms baseline methods.
\end{abstract}

\section{Introduction}
\label{introduction}

Data acquisition, or simply measurement, is one of the most important parts in science and engineering. However, measurement has many of natural corruptions to acquire complete data such as, mainly, additive noise and missingness. Without using corrupted raw data, in the field of signal process, various methods have been studied to retrieve complete data, e.g., denoising for additive noise and imputation for missingness. Historically, Wiener filer \cite{helstrom1967image}, wavelet transform \cite{donoho1995noising} and singular value decomposition \cite{brand2002incremental} are used to denoise image or audio signal. On the other hand, in the field of physics or system design, there have been much more studies to tackle the natural corruption in the sense of making sensors better. While the field of signal processing has utilized machine learning frequently \cite{elad2006image,burger2012image,zhang2017beyond}, the other fields have focused on hardware design and low-level data processing rather than machine learning techniques.
\setlength{\textfloatsep}{15pt}
\begin{figure}[t]
\vskip 0.2in
\begin{center}
\centerline{\includegraphics[width=\columnwidth]{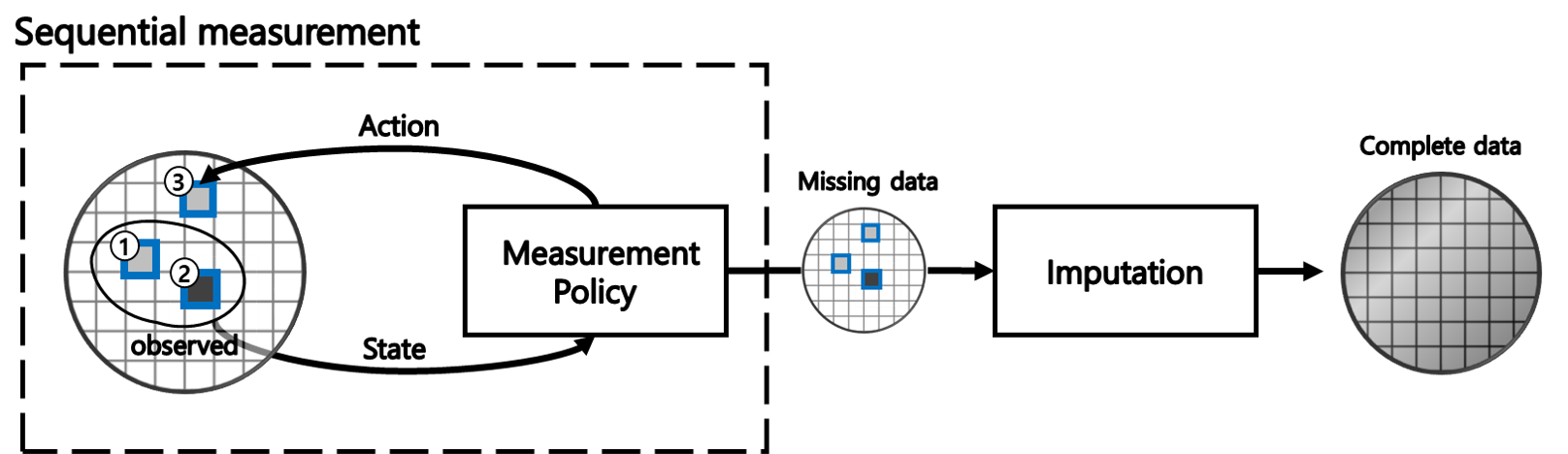}}
\caption{In sequential measurement, policy chooses next variable to be measured for given observed variables. After the iteration stops incompletely, resulting missing data is imputed into complete data.}
\end{center}
\vskip -0.2in
\end{figure}

Missingness, one of the major natures of measurement, can occur in various ways. A stereotype of missingness is caused by measurement failure, i.e., someone tries to measure but cannot. Otherwise, another type of missingness is caused by measurement policy, i.e., someone decides not to measure. We can call the second type of missingness as controllable missingness. Generally speaking, controllable missingness needs an optimal measurement policy to maximize measurement performance constrained by the cost and limitations. For example, in factory manufacturing process, there exist various measurement policies from static ones like uniform sampling to adaptive ones concerning current state of manufacturing process \cite{nduhura2013literature}. Also, sampling techniques used in big data analysis can be interpreted as the measurement policy \cite{settles2009active,rojas2017sampling}. In this paper, we formulate a machine learning problem to find optimal measurement policy based on imputation.

One of the major topics about missingness is imputation to retrieve complete data from missing data. Since missingness frequently occurs in medical or industrial data, imputation is studied in such fields \cite{barnard1999applications,lakshminarayan1999imputation,mackinnon2010use}. However, imputation may not be the most critical part for the missingness caused by measurement failure. The more important point of such missingness is detection and itself can produce other clues for the following data analysis. Besides, provided that measurement system is sufficiently reliable, corresponding missing rate would be sufficiently low, i.e., highly missing data implies catastrophic measurement failure. Conversely, controllable missingness caused by  measurement policy has different characteristics. As measurement is rejected typically due to cost or interference of measurement, many measurement systems would suffer from the missingness, even if the measurement system is reliable. For example, the missing rate of measurement in semiconductor manufacturing process is very high, because a wafer cannot be sampled frequently due to the cost and interference of the measurement process \cite{lee2019data,yugma2015integration}. Thus, in the controllable missingness with high missing rate, imputation becomes much more critical and challenging. Especially, when uncertainties of missing values are too high to be represented uni-modal distribution, multiple imputation is necessary \cite{buuren2010mice, yoon2020gamin}.

In this paper, therefore, we focus on the controllable missingness driven by measurement policy but the acquired data is highly missing so as to be imputed well. To this end, we formulate a reinforcement learning problem for measurement policy to maximize the resulting imputation performance. Also, we will discuss about the detail of scenario which is the most reasonable but challenging to tackle with. First of all, 1) We assume that we can observe the data components sequentially. If we should observe data at once without any prior observation, there is no need to make policy. Thus, in our scenario, policy repeats observation and decision recursively, i.e., we decide to observe from previous observations. Also, 2) We assume that we cannot observe any complete data for learning. In some cases, there would be hard frequency limitation for measurement which disallows us to compute exact imputation performance to be maximized. For the purpose of generality, 3) we assume that there is no prior knowledge for data. More precisely, as we do not have any measurement policy except most uninformative one, the corresponding missingness would be uniformly missing at completely random (MCAR). Note that this does not imply that we focus only on the case of MCAR, because the missingness driven by learned measurement policy can be the case of missing at random, as we discuss in Section \ref{Analysis}. In short, our proposal is to find optimal controllable missingness by learning measurement policy from uncontrollable missingness driven by uninformative measurement policy.

To this end, we propose simple but efficient learning algorithm for both measurement policy and imputation. Our learning algorithm train both jointly despite of the absence of complete dataset. While imputation can be learned without complete dataset in various ways \cite{lakshminarayan1996imputation,raja2020missing,yoon2018gain}, reinforcement learning for measurement policy requires complete data mainly for two reasons. Without complete data, reward for imputation performance should be replaced with less accurate one. Also, more severe one is that measurement policy cannot even explore any action space except the initial uninformative actions. To tackle this problem, we propose a general complete data generation method from the idea that imputation implies conditional generation of complete data. Another important point of the reinforcement learning for measurement policy is that the imputation actually belongs to environment but it can response Unintentionally. Thus, imputation needs to adapt for possible missingness driven by learned measurement policy. 

In proposed learning algorithm, the measurement policy and the imputation are jointly learned. Imputation is learned not only in the sense of adaptation to measurement policy, as mentioned above, but also in the sense of continual or incremental learning. Imputation should learn more from potentially better measurement policy, while it plays a role of complete data generation also. On the other hand, measurement policy is also learned by two perspectives. The first one is to learn better measurement policy without adaptation of imputation and the second one is to learn potentially better measurement policy which leads adaptation positively.

In this paper, we introduce related works in Section \ref{Related Works}, formulated our problem in Section \ref{Problem Statement}, analyze the problem generally in Section \ref{Analysis}, propose the detail of our learning algorithm in Section \ref{Methods}, show implementations and experimental results for sinusoidal dataset and MNIST \cite{lecun1998gradient} in Section \ref{Experiments}, and conclude in Section \ref{Discussion}.

\section{Related Works}
\label{Related Works}

In this section, we introduce several works from general definitions to contemporary studies related with our proposal. Especially, we introduce important concepts of missingness and imputation with respect to them. Plus, we shortly about fundamentals of reinforcement learning and actor-critic method which we mainly use at implementation.

\subsection{Missingness}
Typically, missingness can be represented with binary mask $\boldsymbol{\mathrm{m}}$ of which component is 1 if observed, and 0 if not observed. Then, in order to describe the statistics of missingness, the distribution of mask is utilized. There are three types of missingness in terms of dependency between mask and data. The first type is missing completely at random, call as MCAR, of which mask is independent with entire data. Since both observed data and unobserved data are independent with the missingness, data analysis is unbiased regardless of imputation. The second type is missing at random, called as MAR, of which mask is independent with unobserved data. This type of missingness typically relies on assumption that most of the important data dependent with mask could be observed well. The third type is missing not at random, MNAR, of which mask is also dependent with unobserved data. In short, we can represent those types of missingness using the distribution of mask as below:
\begin{equation}
\label{eqn:1}
P(\boldsymbol{\mathrm{m}}|\boldsymbol{\mathrm{x}})=
  \begin{cases}
    P(\boldsymbol{\mathrm{m}})       & \quad \text{if MCAR} \\
    P(\boldsymbol{\mathrm{m}}|\boldsymbol{\mathrm{x}}_{\text{o}})       & \quad \text{if MAR} \\
    P(\boldsymbol{\mathrm{m}}|\boldsymbol{\mathrm{x}}_{\text{o}}, \boldsymbol{\mathrm{x}}_{\setminus\text{o}})       & \quad \text{if MNAR}
  \end{cases},
\end{equation}
where $\boldsymbol{\mathrm{x}}$ is complete data, $\boldsymbol{\mathrm{m}}$ is mask for missing data, $\boldsymbol{\mathrm{x}}_{\text{o}}$ is observed data for the mask, and $\boldsymbol{\mathrm{x}}_{\setminus\text{o}}$ is unobserved data. Note that we refer $\boldsymbol{\mathrm{x}}_{\text{m}}$ as missing data of which observed components are filled with observed data and unobserved components are filled with NaN, not a number to represent unobserved components explicitly. This can be represented as below:
\begin{equation}
\label{eqn:2}
\boldsymbol{\mathrm{x}}_{\text{m}}^{i}=
  \begin{cases}
    \boldsymbol{\mathrm{x}}^{i}       & \quad \text{if } \boldsymbol{\mathrm{m}}^{i}=1\\
    \text{NaN}       & \quad \text{if } \boldsymbol{\mathrm{m}}^{i}=0
  \end{cases},
\end{equation}
where $\boldsymbol{\mathrm{x}}^{i}_{\text{m}}$, $\boldsymbol{\mathrm{x}}^{i}$ and $\boldsymbol{\mathrm{m}}^{i}$ are $i$-th component of $\boldsymbol{\mathrm{x}}_{\text{m}}$, $\boldsymbol{\mathrm{x}}$ and $\boldsymbol{\mathrm{m}}$, respectively. It is true that this classification of missingness is important for complete data generation and imputation. Thus, we discuss the relationship between this classification and measurement policy in Section \ref{Analysis}.

\subsection{Imputation}
As if missingness is represented with mask, imputation can be represented with masking function for missing data which replaces missing values with other feasible values. Since missingness does not include corruption of observed components by definition, most of the imputation methods preserve observed components. It can be represented with a masking function $M$ and a substitution $\boldsymbol{\mathrm{y}}$ as below:
\begin{equation}
\label{eqn:3}
M(\boldsymbol{\mathrm{x}}_{\text{m}},\boldsymbol{\mathrm{m}},\boldsymbol{\mathrm{y}})^{i}=
  \begin{cases}
    \boldsymbol{\mathrm{x}}^{i}       & \quad \text{if } \boldsymbol{\mathrm{m}}^{i}=1\\
    \boldsymbol{\mathrm{y}}^{i}       & \quad \text{if } \boldsymbol{\mathrm{m}}^{i}=0
  \end{cases},
\end{equation}
where $M(\boldsymbol{\mathrm{x}}_{\text{m}}, m, \boldsymbol{\mathrm{y}})^{i}$ and $\boldsymbol{\mathrm{y}}^{i}$ is $i$-th component of $M(\boldsymbol{\mathrm{x}}_{\text{m}}, m, \boldsymbol{\mathrm{y}})$ and $\boldsymbol{\mathrm{y}}$, respectively. Note that this substitution step is performed not only at the final stage of imputation but also at the other steps or even initial step to for the very first computation. For example, to process missing data with neural network, symbolic missing value should be replaced with a numeric number. Generally, imputation methods are mainly about how to find feasible substitution. One of the most significant way is to regress the substitution from observed data \cite{gelman2006data,zhang2011shell}. The regression using data statistics can be extended to learning problem to extract high-level features from observed data. Especially, as complete data is hard to be observed, unsupervised learning from missing dataset is widely studied \cite{carreira2011manifold,wei2018flexible}. Recently, several unsupervised learning methods using deep neural network show top results for high dimensional and highly missing dataset \cite{yoon2018gain,li2019misgan}. For those challenging dataset, the uncertainty of missing values is too high to be represented with uni-modality. Instead, multiple imputation method is utilized to represent and learn the uncertainty \cite{buuren2010mice,yoon2020gamin}.

\subsection{Reinforcement learning}
Fundamental model of reinforcement learning is modeled as Markov Decision Process composed of agent and environment transmitting stochastic state, action and reward \cite{sutton2018reinforcement}. At each iteration, agent takes action by recognizing current state, and then state is updated from the previous state and the action taken. Agent also receives reward to maximize expected cumulative reward by changing its own policy. There are many reinforcement learning algorithms based on Monte Carlo method, temporal difference, Q-learning and policy gradient method. Among the policy gradient methods, actor-critic method \cite{konda2000actor,mnih2016asynchronous} is widely used from simple scenarios to complicated ones. These days, reinforcement learning works well for some challenging scenarios, in terms of deep reinforcement learning \cite{silver2016mastering,brown2017libratus,levine2016end,pan2017virtual,deng2016deep,franccois2018introduction}. Among them, the most similar method with our proposal is to decide which image patch needs to be obtained with higher resolution \cite{uzkent2020learning}. However, all of the image patches are initially obtained with low resolution and selected at once.

\section{Problem Statement}
\label{Problem Statement}

In this section, we state our problem in detail. Generally speaking, we want to formulate a learning problem to measure data well against missingness. However, we focus on controllable missingness driven by measurement policy rather than accidental missingness caused by measurement failure. From such perspective, we define our learning problem including reasonable dataset acquisition procedure.

First of all, we assume that our measurement system can measure sequentially. At each measurement, measurement policy decides next measurement from already observed data. For simplicity, measurement policy measures one variable per decision and does not memory the history except previous one. Using the mask representation for missingness, stochastic measurement policy can be represented as below:
\begin{equation}
\label{eqn:4}
\boldsymbol{\mathrm{m}}^{(t+1)}{\sim}P(\boldsymbol{\mathrm{m}}|\boldsymbol{\mathrm{m}}^{(t)},\boldsymbol{\mathrm{x}}_{\text{o}}^{(t)}),
\end{equation}
where $\boldsymbol{\mathrm{m}}^{(t)}$ and $\boldsymbol{\mathrm{x}}_{\text{o}}^{(t)}$ are mask and observed data at discrete time $t$, respectively. Also, the distribution $P$ has a constraint that new decision do not overlap with past measurements. Thus, the sum of probabilities of possible masks which overlap all of the past measurements is one as following:
\begin{equation}
\label{eqn:5}
\sum_{\boldsymbol{\mathrm{m}}{\in}S(\boldsymbol{\mathrm{m}^{(t)}})} P(\boldsymbol{\mathrm{m}}|\boldsymbol{\mathrm{m}}^{(t)},\boldsymbol{\mathrm{x}}_{\text{o}}^{(t)})=1.
\end{equation}
Also, to simplify the measurement limitation, we set a maximum length of measurement sequence $T$. Otherwise, actual cost and interference can be modeled. 
\begin{equation}
\label{eqn:6}
S(\boldsymbol{\mathrm{m}^{(t)}})=\{\boldsymbol{\mathrm{m}}^{(t)}+\boldsymbol{\mathrm{m}}\mid \boldsymbol{\mathrm{m}}^{(t)}*\boldsymbol{\mathrm{m}}=\boldsymbol{0}, ||\boldsymbol{\mathrm{m}}||=1\}
\end{equation}
Secondly, we need to define performance metric. While there might be specific goal to be analyzed, retrieving complete data from missing data equivalent to imputation can be general goal before specific analysis. Thus, our performance metric is simply imputation performance and the resulting optimality is also defined by imputation performance. However, our problem is not only to find optimal measurement policy but also to find corresponding optimal imputation method. Thus, joint learning problem for both measurement decision and imputation method can be as follows:
\begin{equation}
\label{eqn:8}
\argmax_{p_{im},\pi} {\mathbb{E}_{\boldsymbol{\mathrm{x}}\sim{}p(\boldsymbol{\mathrm{x}})}[p_{\text{im}}(\boldsymbol{\mathrm{x}}|\boldsymbol{\mathrm{x}}_{o}^{\pi})]},
\end{equation}
where $\boldsymbol{\mathrm{x}}_{o}^{\pi}$ is the data observed by measurement policy ${\pi}$, $p_{\text{im}}$ is distribution of imputation and $p$ is true distribution of complete data. The two individual learning procedures are quite different and there might be no guarantee that alternating the two procedures can generally find global or even local optima. Especially, reinforcement learning should involve exploration resulting in less feasible measurement. Thus, we assume that imputer is well-trained initially by uninformative measurement policy so that imputer can provide reliable reward for exploration at learning measurement policy. Then, provided that measurement policy can be learned monotonically and make missingness somehow easier, imputation also could be learned stably without betraying the measurement policy. In addition to these assumptions, our learning algorithm considers to stabilize the learning in various ways in terms of data generation and meta-learning structure.

On the other hand, since we assume that any complete data cannot be observed, dataset for learning should be missing dataset or even highly missing dataset. One of the most reasonable scenario is uniformly randomly missing dataset which is equivalent to a scenario of uninformative measurement policy. If we have no idea about measurement system, we can just try to measure uniformly randomly selected variables. However, we cannot compute value function in Eq. \ref{eqn:8} without complete dataset. Our proposed algorithm reduces the original problem into another solvable problem using complete data generation.

\section{Analysis}
\label{Analysis}

In this section, we discuss about specific characteristics of our problem. Even though it is not theoretically proved, some of the intuitions can inform how to solve problem and why it can be solved.

First, we can think about the existence of optimal measurement policy. The fundamental reason why we can regress imputation is because each variables of complete data are correlated. Then, which unobserved variable should be measured for given observed data? It can be said that the most desired unobserved variable should be correlated mostly with the other unobserved variables which are less correlated with observed variables. However, how can we know about the correlation between observed variables and unobserved ones. A trivial way is to check global correlation between them. For example, in MNIST, reliable guess would be on a pixel around center but in distance from observed pixels. Nevertheless, we need to make more reliable guess for given observed variables. If there exist multiple hypotheses on correlation for given observed variables, we can measure a variable to test the hypothesis and discard if it is unreliable. An explicit form of hypothesis on correlation can be that on complete data. Thus, the most desired variable is the most uncertain variable among various hypotheses on complete data from multiple imputation. Therefore, at least, learning measurement policy should involve implicit learning of hypotheses. Hopefully, measurement policy learned by imputation can learn implicit hypotheses but more optimal ones to decide which variable is important. An important type of implicit hypotheses of correlation should come from imputation method itself. Since any imputation method could be somehow overfitted, measurement decision rule can reduce generalization error possibly caused by the particular imputation method.

On the other hand, imputation should be learned jointly with measurement policy. It has two major reasons: The first one is to adapt to missingness changed by learning measurement policy, and second one is to learn more from potentially better missingness. Considering the adaptation, the dependency of missingness upon the data changes through learning. Initially, the missing data driven by uninformative measurement policy is MCAR. Then, after missingness is controlled by observed data, it belongs to MAR case. While the adaption is ambiguous for contribution at imputation, it is obvious that better measurement policy make imputation better. Practically, most imputation methods practically do better at easier missing data without additional learning. Even though fixed imputation method cannot do better at easier missingness, we can hopefully believe that a proper learning method can make imputation better.

As we can expect the desired behaviors of proper learning algorithm and optimal policy as above, it is different from implementability in the perspective of function approximation. Initially, a function approximating measurement should be nearly irrelevant with input, because it should represent MCAR missingness. After learning, the function should be quite sensitive to response for slowly changing input. Especially, the score for previously selected component should extremely fall down, unless the score is rejected in hard-coded way. Thus, function and input need to be designed carefully rather than typical function approximation methods in machine learning which desires robust and smooth function approximation.

\section{Methods}
\label{Methods}

\begin{figure}[t]
\vskip 0.2in
\begin{center}
\centerline{\includegraphics[width=\columnwidth]{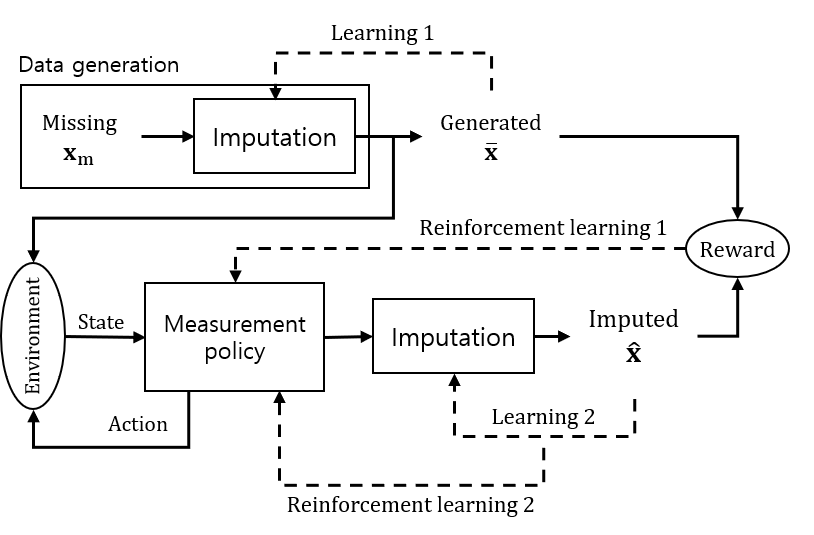}}
\caption{Above learning procedures are scheduled for joint learning of measurement policy and imputation. 'Learning 1' is for data generation based on the imputation and 'Learning 2' is for adapting the imputation for the measurement policy. 'Reinforcement learning 1' receives the reward from imputation and 'Reinforcement learning 2' receives the reward from the adapted imputation by 'Learning 2'.}
\label{LEARNING_PROCEDURE}
\end{center}
\vskip -0.2in
\end{figure}

In this section, we propose a joint learning algorithm for both measurement policy and imputation. The intuition of the learning procedure is in Figure \ref{LEARNING_PROCEDURE}. Also, it is based on deep learning including function approximation using neural networks and stochastic gradient descent. First of all, we let the state for measurement policy be re-defined missing data representation as below:
\begin{equation}
 \label{equation100}
 \boldsymbol{\mathrm{x}}_{\text{m}}^{(t)}=[\boldsymbol{\mathrm{x}}^{(t)}_{\text{o}},\boldsymbol{\mathrm{m}}^{(t)}],
\end{equation}
where $\boldsymbol{\mathrm{x}}_{\text{m}}^{(t)}$ is missing data at discrete time $t$ re-defined as a pair of observed data and mask.
\subsection{Initialization}
As we assume that our dataset is uniform MCAR dataset driven by the uninformative measurement policy, we need to initialize measurement policy network to represent the distribution. While an initialized neural network easily produces uniform distribution, it can hardly produce irrelevant output. Thus, we use dropout layers \cite{hinton2012improving} to inject additional randomness to neural network and learn dependency through training. Since the initialized measurement policy network is not completely uniform MCAR, we need to use a sampling technique for the uninformative measurement policy. Then, imputation is pre-trained by sampled missing dataset to prevent the measurement policy network from being confused by unreliable imputation. Fortunately, we can train imputation with unsupervised manner using only missing dataset \cite{yoon2018gain, li2019misgan, yoon2020gamin}.

\subsection{Data generation}
 
Unfortunately, we cannot explore measurement policies even by clever exploration method, because none of the measurement is included except the observed one in a given missing dataset. Furthermore, our reinforcement learning algorithm cannot compute the reward defined by actual imputation performance without complete dataset. To tackle these impossibilities, we propose a simple but efficient complete data generation method. The intuition is that imputation belongs to conditional generation of complete data from missing data. The only difference is whether each generation is constrained by true distribution or other desired distributions. Thus, the complete data generated by imputation is the most realistic one among data generation methods. Additionally, using stochastic or multiple imputation, we can make larger dataset. In such case, how to generate complete dataset optimally also belongs to machine learning. An active learning method can possibly evaluate imputation quality of each data. Without any active sampling , a complete data $\bar{\boldsymbol{\mathrm{x}}}$ is generated as below:
\begin{equation}
\label{equation105}
\bar{\boldsymbol{\mathrm{x}}} \sim{} p_{\text{im}}(\boldsymbol{\mathrm{x}} \vert \boldsymbol{\mathrm{X}}_{\text{m}} ; \phi),
\end{equation}
where $p_{\text{im}}$ is imputation distribution parameterized with $\phi$.

Then, the individual learning measurement policy and imputation can be realized by the data generation method. Firstly, the reinforcement learning algorithm can explore any policies on the generated complete data. Also, the reward function can compute imputation error between imputation from the measurement policy and its corresponding complete data. In entire episodes, it would be sufficient to use only terminal reward as final imputation error rather than immediate rewards from every single measurement. Thus, the gradient driven by REINFORCE algorithm \cite{sutton2018reinforcement} is as below:
\begin{equation}
\label{equation104}
\nabla_{\theta}L(\boldsymbol{\mathrm{x}})= {\mathbb E}[R^{(t)}\nabla_{\theta} \log  \pi(\boldsymbol{\mathrm{m}}^{(t)}\vert\boldsymbol{\mathrm{x}}^{(t)}_{\text{m}};\theta )],
\end{equation}
where $R^{(t)}$ is the cumulative reward at time $t$ computed with the generated complete data $\boldsymbol{\mathrm{x}}$. Also, since we assume that one variable is observed at each time, the policy $\pi$ is defined as below:
\begin{equation}
\label{equation101}
\pi(\boldsymbol{\mathrm{m}}\vert\boldsymbol{\mathrm{x}}^{(t)}_{\text{m}};\theta )= \text{Cat}\left[ \frac{(\boldsymbol{1}-\boldsymbol{\mathrm{m}}^{(t)})e^{f(\boldsymbol{\mathrm{x}}^{(t)}_{\text{m}};\theta)}}{\sum (\boldsymbol{1}-\boldsymbol{\mathrm{m}}^{(t)})e^{f(\boldsymbol{\mathrm{x}}^{(t)}_{\text{m}};\theta)}}\right],
\end{equation}
where $\pi$ is a stochastic policy represented by a categorical distribution and $f(\bigcdot;\theta)$ is a policy network parameterized by $\theta$. Furthermore, for exploration, we flatten the distribution slightly as below:
\begin{equation}
\label{equation103}
\pi_{e} = (1-e)\pi + e(1-\pi),
\end{equation}
where $e\in[0,1]$ is a hyper-parameter to control the magnitude of exploration. Note that the flattened distribution is invalid so as to be normalized.

On the other hand, the learning method equivalent to that of initialization using real missing dataset is available, a supervision using generated complete data is also available. However, it is important to maintain imputation performance on the real missing data, because it is used to generate complete data affecting learning recursively. Thus, the gradients can be written as:
\begin{equation}
\label{equation106}
\nabla_{\phi}L_{\text{im}}^{u}(\boldsymbol{\mathrm{x}}_{\text{m}})\;\text{or}\;\nabla_{\phi}L_{\text{im}}^{s}(\boldsymbol{\mathrm{x}}_{\text{m}}, \boldsymbol{\mathrm{x}}),
\end{equation}
where $L_{\text{im}}^{u}$ and $L_{\text{im}}^{s}$ are loss functions based on unsupervised and supervised learning, respectively.

In the case of learning from the better measurement policy learned as above, there are two possible approaches for adaptation or additional learning. The first one is to use a new imputation method more appropriate to easier missingness and transfer the knowledge from the original imputation method. The second approach is to share identical imputation method for both data generation and imputation to take the benefit of data augmentation.
\setlength{\textfloatsep}{15pt}

\subsection{Joint learning}
 
There needs to be several techniques to make joint learning possible and stable. Simply, we want to replace the joint learning with an alternate learning algorithm in the sense of inner loop. In addition to computational efficiency, hopefully, it can be better than the alternating individual learning in the sense of outer loop. Firstly, the progressive measurement which occurs due to exploration of reinforcement learning can perturb imputation. Thus, imputation should adapt to the current stationary policy. Also, we adopt an additional learning procedure similar to meta-learning. By definition, a potentially better measurement policy should encourage imputation to adapt in positive way. Thus, the reward function for measurement policy should consider the imputation performance after adaption. To this end, we add another reward which is imputation error after an update of imputation, which is computationally reasonable for iteration. The progressive adaptation resulted by exploring the influence onto adaptation is discarded and the actual adaptation occurs after the update of the measurement policy. These procedures are described in Algorithm \ref{alg:1}.
\setlength{\textfloatsep}{15pt}
\begin{algorithm}[t]
\caption{Joint learning}
\label{alg:1}
\begin{algorithmic}
    \setstretch{1.2}
    \REQUIRE measurement policy $\pi$, imputation $p_{\text{im}}$
    \STATE {Input:} policy parameter $\theta$, imputation parameter $\phi$, missing dataset $\boldsymbol{\mathrm{X}}_\text{m}$
    \STATE {hyperparameters:} $\alpha$, $\alpha'$, $\beta$, $\beta'$, $e$ \\[7pt]
    \STATE Initialize $\phi$
    \REPEAT
    \STATE Sample batch of missing data $\boldsymbol{\mathrm{x}}_\text{m}$ from  $\boldsymbol{\mathrm{X}}_\text{m}$
    
    \STATE $\phi\gets\phi-\alpha\nabla_\phi L_{\text{im}}\left(\boldsymbol{\mathrm{x}}_\text{m};p_{\text{im}}\right)$
    
    \UNTIL{$L_\text{im}\left(\boldsymbol{\mathrm{x}}_\text{m}\right)$ converges}\\[7pt]
   \STATE Initialize $\theta$
    \REPEAT
    \setstretch{1.3} 
    \STATE Sample batch of imputation $\boldsymbol{\mathrm{\bar{x}}}$ from $p_{\text{im}}\left(\bigcdot\mid\boldsymbol{\mathrm{x}}_\text{m};\phi\right)$
    
    \STATE Exploring policy $\pi_e\left( \bigcdot;\theta \right)\gets (1-e)\pi + e(1-\pi)$
    
    \STATE Sample episode $E_1=\{ {\boldsymbol{\mathrm{\bar{x}}}_\text{m}^{(t)}} \}_{t=1}^{t=T}$ using $\pi_e\left( \bigcdot;\theta \right)$
    
    \STATE $\phi_{\text{new}}\gets\phi - \alpha\nabla_\phi L_{\text{im}}\left(\boldsymbol{\mathrm{x}}_\text{m};p_{\text{im}}\right) - \alpha'\nabla_\phi L'_{\text{im}}(\boldsymbol{\mathrm{\bar{x}}}_\text{m}^{(T)};p_{\text{im}})$
 
    \STATE Sample episode $E_2=\{\boldsymbol{\mathrm{\bar{x}}'}_\text{m}^{(t)}\}_{t=1}^{t=T}$ using $\pi\left( \bigcdot;\theta\right)$

    \STATE $\theta\gets\theta-\beta\nabla_\theta L\left(\boldsymbol{\mathrm{\bar{x}}},  E_1,  p_{\text{im}}\left( \bigcdot;\phi\right)\right)\newline\hspace*{0.9cm}-\beta'\nabla_\theta L\left(\boldsymbol{\mathrm{\bar{x}}}, E_2,  p_{\text{im}}\left( \bigcdot;\phi_{\text{new}}\right)\right)$
   
    \STATE Sample episode $E_3=\{{\boldsymbol{\mathrm{\bar{x}}''}_\text{m}^{(t)}}\}_{t=1}^{t=T}$ using $\pi\left( \bigcdot;\theta\right)$
   
    \STATE $\phi\gets\phi-\alpha\nabla_\phi L_{\text{im}}\left(\boldsymbol{\mathrm{x}}_\text{m};p_{\text{im}}\right)-\alpha'\nabla_\phi L'_{\text{im}}(\boldsymbol{\mathrm{\bar{x}}''}_\text{m}^{(T)};p_{\text{im}})$
    \UNTIL{convergence}
\end{algorithmic}
\end{algorithm}

\section{Experiments}
\label{Experiments}
In this section, we evaluate our proposal in two different datasets: MNIST dataset for complex scenario and sinusoidal dataset for simple scenario. We demonstrate the implementation details, results with various missing rates and ablation studies. More details such hyperparmeters are described in Supplementary material.

\subsection{MNIST dataset}

We implement the measurement system and learning algorithm using MNIST dataset \cite{lecun1998gradient}. For imputation, we use generative adversarial network structure \cite{yoon2020gamin} which can learn multiple imputation in unsupervised manner. The structure consists of unconditional generator for candidate substitution and conditional generator for imputation with respective discriminators. However, for adaptation, we only trained the generator for imputation and corresponding imputation, because the candidate generation is based on MCAR assumption. For measurement policy, we used actor-critic method for REINFORCE method \cite{sutton2018reinforcement}. The reward function is defined as top-$k$ RMSE in \cite{yoon2020gamin} as below: 
\begin{equation}
\label{terminal_reward}
R_{k}=\min_{i<k}\left\Vert{\hat{\boldsymbol{\mathrm{x}}}_{i} -\boldsymbol{\mathrm{\bar{x}}}}\right\Vert_2,
\end{equation}
where $\hat{\boldsymbol{\mathrm{x}}}_{i}$ is top-$i$ imputation and $\boldsymbol{\mathrm{\bar{x}}}$ is generated complete data. Also, we utilize top-3 RMSE using ground truth $\boldsymbol{\mathrm{x}}$ for evaluation metric.

We use 60,000 examples for training and 10,000 for test as provided in MNIST dataset. For simplicity, we crop and resize the image into 12x12 resolution. Also, we use fully connected network for every network including generators, discriminators, actor and critic. In the supplementary material, we demonstrate experimental results of original resolution 28x28.

\setlength{\textfloatsep}{15pt}
\begin{figure}[h]
\begin{center}
\vskip 0.2in
\centerline{\includegraphics[width=\columnwidth]{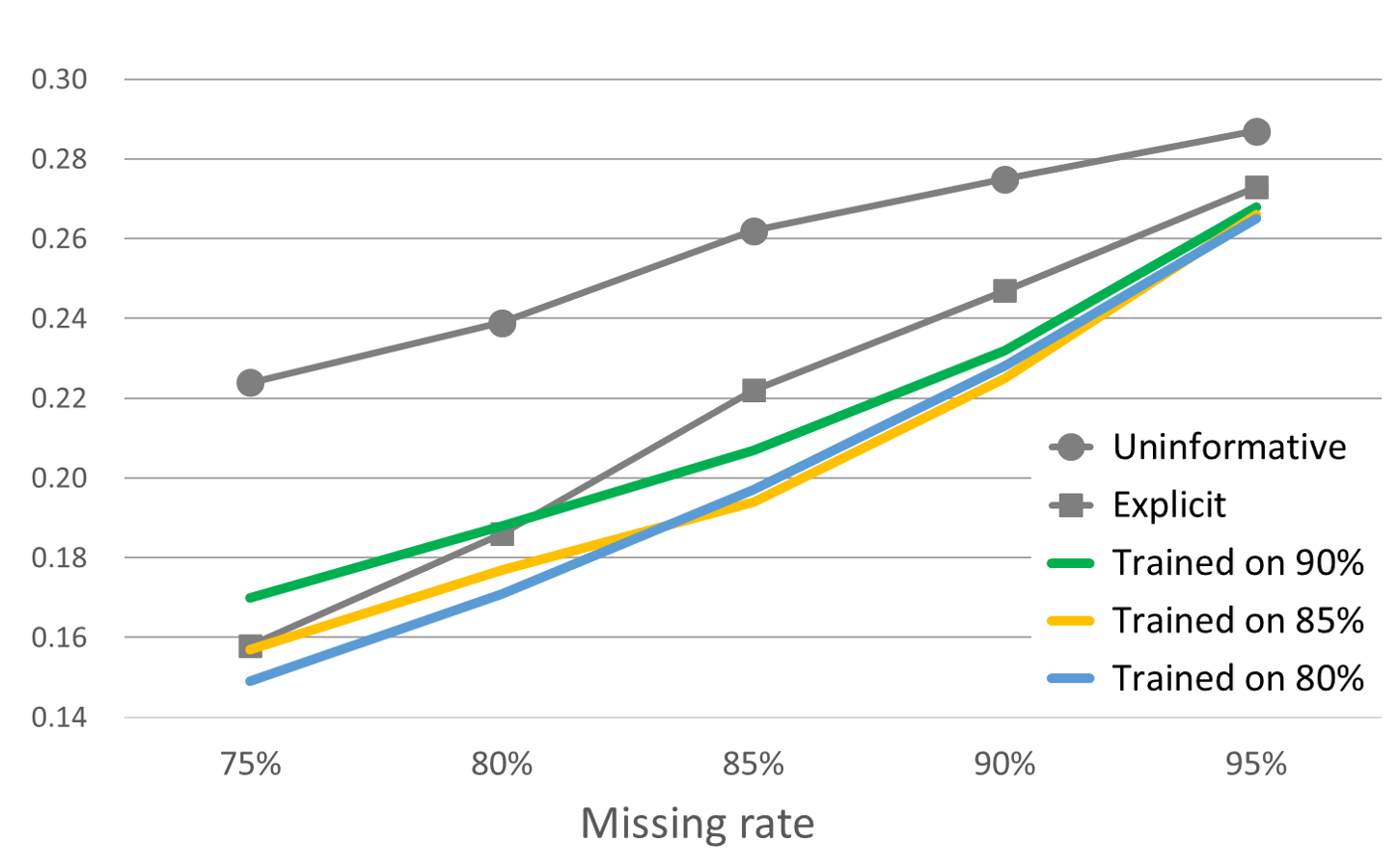}}
\vskip -0.2in
\end{center}
\caption{MNIST dataset: Top-3 imputation errors of baseline methods and our proposal trained on various missing rates.}
\label{MNISTRESULT}
\end{figure}

\setlength{\textfloatsep}{15pt}
\begin{figure}[h]
\begin{center}
\vskip 0.2in
\centerline{\includegraphics[width=\columnwidth]{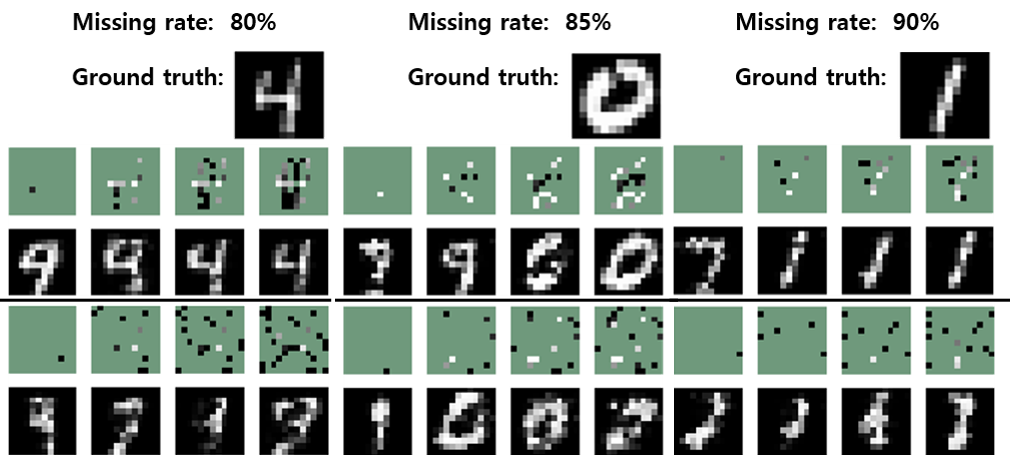}}
\vskip -0.2in
\end{center}
\caption{Examples of MNIST dataset: The top two rows are from proposed method and the bottom two rows are from uninformative method}
\label{MNIST_VISUALIZATION}
\end{figure}

\setlength{\textfloatsep}{15pt}
\begin{table}[h]
\begin{center}
    \begin{tabular}{>{\centering}m{1.8cm} >{\centering}m{1.52cm} >{\centering}m{1.52cm} >{\centering}m{1.52cm}}
    \Xhline{3\arrayrulewidth}
    Method & 80\% & 85\% & 90\% \tabularnewline
    \Xhline{3\arrayrulewidth}
    UnInform & .239 (.260) & .262 (.284) & .275 (.302) \tabularnewline
    \hline
    Explicit & .186 (.199) & .222 (.238) & .247 (.265) \tabularnewline
    \hline
    Proposed & .171 (.185) & .194 (.213) & .232 (.251) \tabularnewline
    \Xhline{3\arrayrulewidth}
\end{tabular}
\end{center}
\caption{MNIST dataset: Top-3 imputation errors of baseline method and our proposal trained in various missing rate. Single imputation error is in parenthesis.}
\label{table1}
\end{table}

Firstly, we evaluate for varying missing rate in Table \ref{table1}. 'UnInform' is uninformative measurement policy and corresponding imputation method. 'Explicit' is an explicit policy choosing the most uncertain variable from multiple imputations at every measurement. 'Proposed' is our measurement policy jointly learned with imputation and deterministic policy is evaluated. Table \ref{table1} shows imputation performance of each policies and corresponding imputation. For each missing rate, imputations of baseline methods and initial imputations of our learning algorithm are identically trained at the missing rate and the measurement policy observes the same number with training data. While explicit policy shows remarkable results compared to uninformative one, our proposal shows better performance, as expected. Also, Figure \ref{MNIST_VISUALIZATION} shows examples visualizing each measurement policy and imputation result. Furthermore, we tested each policy at various missing rates different with the missing rate used in training, as shown in Figure \ref{MNISTRESULT}. From the result, generalization on less observation is more reliable than on more observation. Note that the performance of explicit policy is the minimum of training results at the three different missing rate.
\setlength{\textfloatsep}{15pt}
\begin{table}[h!]
\begin{center}
    \begin{tabular}{>{\centering}m{1.5cm} >{\centering}m{0.7cm} >{\centering}m{0.7cm} >{\centering}m{0.7cm} >{\centering}m{0.7cm}>{\centering}m{0.7cm}}
    \Xhline{3\arrayrulewidth}
    Method & 75\% & 80\% & 85\%* & 90\% & 95\% \tabularnewline
    \Xhline{3\arrayrulewidth}
    w/o adaptation & .161 (.175) & .183 (.199) & .203 (.222) & .238 (.260) & .272 (.299) \tabularnewline
    \hline
    w/o meta & .159 (.173) &  .177 (.192) & .199 (.217) & .228 (.250) & .267 (.295) \tabularnewline
    \hline
    Proposed & .157 (.170) & .177 (.193) & .194 (.213) & .225 (.247) & .266 (.295) \tabularnewline
    \Xhline{3\arrayrulewidth}
\end{tabular}
\end{center}
\caption{MNIST dataset: Top-3 imputation errors of three variations of our learning algorithm. Single imputation error is in parenthesis. *All variations are trained on 85\% missing data.}
\label{table2}
\end{table}

Secondly, we investigate the results of three variations trained at 85\% missing rate of our learning algorithm, as shown in Table \ref{table2}. The first one is identical to 'Proposed' in Table \ref{table1} and 'w/o meta' excludes the reward from adapted imputation ($\beta'=0$ in Algorithm \ref{alg:1}). In 'w/o adaptation', policy is individually learned without joint learning with imputation and imputation is fine-tuned later. Note that it also excludes the reward from adaptation like 'w/o meta.' From the result, proposed learning algorithm seems to behave as expected, but the effect on generalization to other missing rate is relatively ambiguous.

\subsection{Sinusoidal dataset}

In order to investigate the performance for simpler dataset, we utilize 1D sinusoidal function $y(x) = A\sin(wx + b)$ parameterized with amplitude $A$, frequency $w$ and phase $b$. Like \cite{finn2017model}, those parameters are sampled uniformly from the range $A\in[0.1, 1]$, $b\in[0, 2\pi]$ and $w\in[0.5, 2]$, respectively. Since the sinusoidal function can be determined with a few observations, it can be said to have uni-modal uncertainty, so we can learn the best single imputation. Thus, the imputation method consists of linear interpolation and simple imputation network. Also, assuming that the complete data is smooth, we adopt smoothness loss function using Gaussian filtering. We sample one hundred variables in regular grid $[-5, 5]$. Also, training dataset has 2,880 examples and test dataset has 720 examples, respectively.
\setlength{\textfloatsep}{15pt}
\begin{table}[h]
\begin{center}
    \begin{tabular}{>{\centering}m{2.3cm} >{\centering}m{1cm} >{\centering}m{1cm} >{\centering}m{1cm}>{\centering}m{1cm}}
    \Xhline{4\arrayrulewidth}
    Method & Single 80\% & Single 90\% & Double 80\% & Double 90\% \tabularnewline
    \Xhline{4\arrayrulewidth}
    UnInform & .036 & .125 & .113 & .264 \tabularnewline
    \hline
    Proposed & .026 & .077 & .096 & .209 \tabularnewline   
    \Xhline{4\arrayrulewidth}
\end{tabular}
\end{center}
\caption{Sinusoidal dataset: Single imputation errors for two types of dataset and missing rate.}
\label{table3}
\end{table}

\begin{figure}[h]
\begin{center}
\vskip 0.2in
\centerline{\includegraphics[width=\columnwidth]{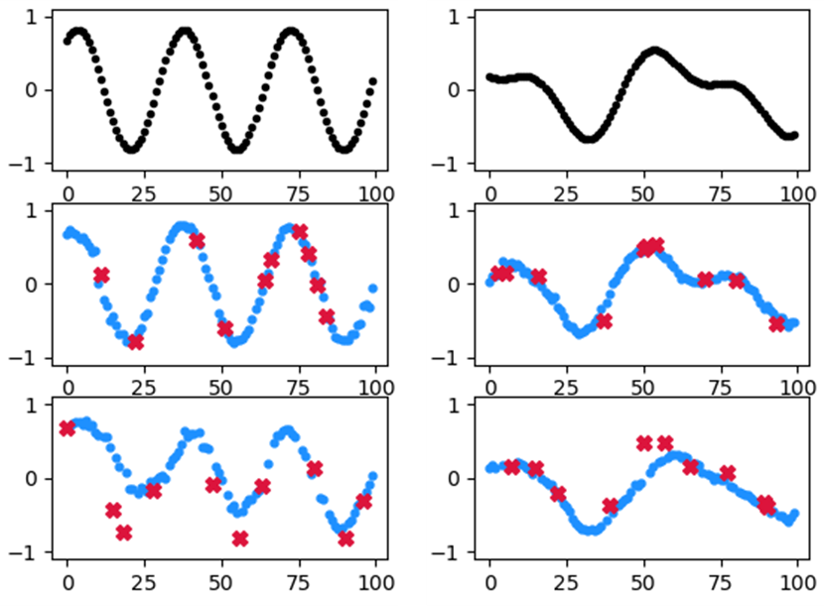}}
\vskip -0.2in
\end{center}
\caption{Examples of sinusoidal dataset: The tops are ground truth, the middles are from proposed method and the bottoms are from uninformative method.}
\label{SIN_VISUALIZATION}
\end{figure}

In Table \ref{table3}, we represent the experimental results for several data characteristics. We utilize stochastic policy for evaluation and the explicit method based on multiple imputation in Table \ref{table1} is not possible. 'Single' is sampled from sinusoidal functions above and 'Double' is sampled by superposition of two different sinusoidal functions, respectively. As missing rate increases, gain from learning measurement policy increases. Figure \ref{SIN_VISUALIZATION} shows each example of 'Single' and 'Double'. Above all, we investigate that our learning algorithm can be applied to simple dataset by learning implicit characteristic other than explicit one.
\setlength{\textfloatsep}{15pt}

\section{Conclusion}
\label{Discussion}
In conclusion, we propose a learning algorithm for both sparse measurement policy and corresponding imputation method. Since we want joint learning due to their mutual relationship, our learning algorithm alternately update them considering clever exploration method. Also, complete data generation method allows learning from only uncontrollable, uninformative missing data. We showed the best results at both simple and complex dataset with various missing rates.

Although we can investigate our algorithm empirically, we cannot derive theoretical analysis except intuitive ones. Interestingly, there might be meaningful properties or theoretical bounds for the applicability of our algorithm. On the other hand, we can extend our learning algorithm for true active learning or interactive learning, assuming that measurement for learning also can be controlled. Practically, however, if measurement should wait those learning procedure which would be not real-time, it also increases another type of cost so that the measurement becomes expensive or even impossible. Thus, more practical future work could be more data-specific or task-specific applications using challenging data characteristics, in addition to theoretical study.

\bibliography{example_paper}
\bibliographystyle{icml2021}

\end{document}